\documentclass[sigconf]{acmart}
\usepackage{listings}

\renewcommand\footnotetextcopyrightpermission[1]{} 
\acmConference[]{}{}{}
\AtBeginDocument{%
  \providecommand\BibTeX{{%
    \normalfont B\kern-0.5em{\scshape i\kern-0.25em b}\kern-0.8em\TeX}}}

\setcopyright{acmlicensed}
\copyrightyear{2024}
\acmYear{2024}
\acmDOI{XXXXXXX.XXXXXXX}
\acmConference[Conference acronym 'XX]{Make sure to enter the correct
  conference title from your rights confirmation email}{June 03--05,
  2018}{Woodstock, NY}
\acmISBN{978-1-4503-XXXX-X/18/06}

\graphicspath{{./images/}} 
\begin{document}

\title{A Framework for Multi-View Multiple Object Tracking using Single-View Multi-Object Trackers on Fish Data}

\author{Chaim Chai Elchik$^{*}$, Fatemeh Karimi Nejadasl$^{*}$, Seyed Sahand Mohammadi Ziabari$^{*}$, Ali Mohammed Mansoor Alsahag$^{*}$}
\affiliation{%
  \institution{$^{*}$Informatics Institute, University of Amsterdam}
  \streetaddress{1098XH Science Park}
  \city{Amsterdam}
  \country{The Netherlands}
}
\email{chaim.elchik@student.uva.nl, f.kariminejadasl@uva.nl, s.s.mohammadiziabari@uva.nl, a.m.m.alsahag@uva.nl}

\begin{abstract}
  Multi-object tracking (MOT) in computer vision has made significant advancements, yet tracking small fish in underwater environments presents unique challenges due to complex 3D motions and data noise. Traditional single-view MOT models often fall short in these settings. This thesis addresses these challenges by adapting state-of-the-art single-view MOT models, FairMOT and YOLOv8, for underwater fish detecting and tracking in ecological studies. The core contribution of this research is the development of a multi-view framework that utilizes stereo video inputs to enhance tracking accuracy and fish behavior pattern recognition. By integrating and evaluating these models on underwater fish video datasets, the study aims to demonstrate significant improvements in precision and reliability compared to single-view approaches. The proposed framework detects fish entities with a relative accuracy of 47\% and employs stereo-matching techniques to produce a novel 3D output, providing a more comprehensive understanding of fish movements and interactions.
\end{abstract}
\maketitle

\section{Introduction}
\label{sec:introduction}
\par Multiple object tracking (MOT) of small elements such as fish presents challenges in effectively modeling object behavior over time, primarily involving two aspects: detection and association. Typically, the association part is not learned, and object ID embeddings are learned separately. These embeddings are then combined using the Hungarian matching algorithm. \cite{zeng2022motr}. The traditional multiple-object tracking problem involves detecting and following multiple objects through a series of frames in a video sequence. This task is challenging due to variations in object appearance, occlusions, and complex interactions between objects \cite{zeng2022motr}.

\par In Multi-Object Tracking (MOT), a single network is often trained to perform both object detection and re-identification tasks simultaneously. This approach, known as multi-task learning \cite{Zhang_2021}, allows for joint optimization but can lead to conflicts between the tasks. Specifically, object detection focuses on accurately identifying all fish in the scene, while re-identification aims to distinguish between individual fish. These conflicting objectives can lead to a situation where improving detection accuracy negatively impacts re-identification performance and vice versa. This inherent conflict between the tasks results in a compromise that can hinder the overall tracking accuracy \cite{Zhang_2021}.

\par The issue revolving around the separation of the tasks within MOT has led to the introduction of FairMOT \cite{Zhang_2021}. FairMOT is based on the anchor-free object detection architecture CenterNet \cite{Zhang_2021}. In FairMOT the detection and re-identification tasks are treated equally instead of using the “detection first, re-ID secondary” framework. The FairMOT framework is also not just a naive combination of CenterNet and re-identification \cite{Zhang_2021}. 

\par Another challenge concerns the temporal modeling of objects. Traditional methods struggle to capture the long-term dependencies between objects across video sequences. This limitation hinders the ability to track objects effectively, especially in scenarios with occlusions or irregular motion patterns \cite{zhang2022bytetrack}.

\par YOLOv8, the latest iteration of the You Only Look Once (YOLO) object detection framework, addresses these challenges by offering improved detection accuracy and efficiency. Its optimized network architecture and advanced training methods allows for faster and more precise object detection in real-time scenarios \cite{ultralyticsHome}. This makes YOLOv8 particularly suitable for MOT applications, as it provides a robust foundation for accurate object tracking.

\par While YOLOv8 excels at object detection, robust tracking requires additional algorithms. ByteTrack offers a solution by refining detections and associating objects effectively. It leverages high-confidence detections from YOLOv8 to guide the tracking of lower-confidence objects. This approach ensures a more stable and accurate tracking process, even in challenging scenarios \cite{zhang2022bytetrack}.

\par By combining YOLOv8's rapid and precise detection capabilities with ByteTrack's robust tracking algorithms, YOLOv8 efficiently identifies objects in each frame, while ByteTrack effectively associates these detections across frames to form consistent object tracks. This synergy overcomes the limitations of separate detection and tracking approaches, leading to superior MOT performance. This approach avoids the pitfalls of traditional "detection first" methods and allows for a more robust and accurate tracking process, especially in real-time applications.
\newline

\par A challenge within 3D object detection is handling scenarios with occlusions and complex object arrangements, especially from a single camera viewpoint. DETR3D addresses this challenge by leveraging multiple cameras to enhance 3D perception \cite{wang2022detr3d}. The model introduces 3D-to-2D queries, enabling it to directly predict 3D object locations from image features. This method emphasizes robust handling of occluded objects and improved accuracy in densely populated scenes \cite{wang2022detr3d}.

\par FairMOT, and Yolov8 (combined with ByteTrack) are both examples of state of the art single single-view multiple object trackers. None of these models or similar models have been adapted or used in the field of ecological studies but due to their nature could theoretically be adapted to become potentially useful tools within this field. The fish data set that will be used is entirely new to the field and unique as these kinds of videos have not been used with these specific models before or any similar models either. This research aims to investigate their potential use for tracking fish movements. The adaption of any of these models then or similar single-view multiply object tracking frameworks into a multi-view multiply tracking framework to try and get even better results is something that has not been attempted before. This approach has the potential to significantly impact the field of multi-view MOT by offering a novel perspective on the problem. Underwater footage is also something new as none of these single-view models have been trained or evaluated on datasets containing underwater footage, something that in itself creates vast challenges. Water uniquely bends light and image quality degradation makes underwater image detection more challenging \cite{Er_2023}. Furthermore, how fish appear from different angles is a challenge, a fish looking straight at the camera appears vastly different from one viewed from the side. The aim of this study is to develop and evaluate a multi-view multiple object tracking (MOT) framework tailored for underwater ecological video data. Building on state-of-the-art single-view MOT models such as FairMOT and YOLOv8 with ByteTrack, the framework is designed to overcome challenges posed by small, visually similar fish and complex 3D motion. Specifically, the research explores how single-view trackers can be leveraged to enable robust fish tracking, support pattern recognition in locomotion, and generate three-dimensional representations through dual video inputs. The integration of a 3D perspective is also assessed in terms of its impact on tracking accuracy and detection reliability in underwater conditions.

\section{Related Work}
\label{sec:related_work}
The research in this proposed thesis builds upon the works of MOT, FairMOT, Yolov8, and ByteTrack which are all advancements in the field of multiple object detection and tracking. It also builds upon multi-view approaches to MOT. The  uniqueness of this thesis lies in its application of MOT, specifically designed for underwater fish tracking and incorporating a multi-view approach through multiple video inputs. None of these MOT models have been used on animal data sets or for ecological studies, nor have any of them been used to create a multi-view approach. This thesis will bring to light a new field in which multi-view multiple object-tracking models can potentially be used successfully. 

\subsection{Single-View Multiple Object Detection and Tracking Models}
Single-view MOT models have been researched extensively which has led to many different models using differing approaches. The section below briefly explains how the chosen and most recent state-of-the-art models work.
\subsubsection{FairMOT}
\par With two homogeneous branches for predicting pixel-wise objectness scores and re-identification features, FairMOT is a multi-object tracking model \cite{Zhang_2021}. High degrees of detection and tracking accuracy are accomplished by utilizing the achieved fairness between the tasks. Position-aware measurement maps are used to describe object centers and sizes, and the detection branch is performed in an anchor-free manner \cite{Zhang_2021}. Similarly, to describe the object centered at each pixel, the re-ID branch estimates a re-ID feature for each pixel. This all made for FairMOT being state of the art for its new approach. Recent work has explored replacing its CNN backbone with Vision Transformers (ViT) for tracking small, occluded targets such as fish ~\cite{anwar2025fairmot}. While promising, ViT did not outperform CNN-based FairMOT models, highlighting challenges in re-ID performance and real-world tracking accuracy.
Recent studies have explored the integration of foundation models to address challenges in ecological video analysis and multi-object tracking. Katona et al.~\cite{katona2025marine} introduced MARINE, a model that combines motion-based frame selection with DINOv2 features to detect rare predator-prey interactions in animal videos, showing strong performance on both domain-specific and general datasets. In parallel, Faber et al.~\cite{faber2024leveragingfoundationmodelsknowledge} applied a knowledge distillation approach, using DINOv2 as a teacher to transfer rich visual representations to a FairMOT-based student model. While the approach demonstrated improvements in certain conditions, it also highlighted limitations in adapting large vision models to lightweight, task-specific architectures. These works collectively underscore both the potential and constraints of using foundation models in real-world and animal-focused tracking scenarios.

\subsubsection{YOLOv8}
\par YOLOv8 builds upon the success of the You Only Look Once (YOLO) object detection series, known for its balance between accuracy and real-time performance \cite{ultralyticsHome}. The core network architecture utilizes efficient building blocks like the Focus module and Path Aggregation Network (PAN) to achieve fast and accurate object detection \cite{ultralyticsHome}. Additionally, YOLOv8 leverages various data augmentation techniques during training, which leads to improved generalization and robustness to variations in real-world data \cite{ultralyticsHome}. These advancements made YOLOv8 state-of-the-art at its time of publication.

\subsubsection{ByteTrack}
\par ByteTrack presents a novel approach to multi-object tracking that departs from traditional methods \cite{zhang2022bytetrack}. Instead of filtering out low-confidence detections, ByteTrack associates all detections with tracklets (short tracks). This approach allows ByteTrack to leverage even uncertain detections, potentially leading to more accurate tracking, especially for objects undergoing motion blur or occlusions \cite{zhang2022bytetrack}. To handle these low-confidence detections, ByteTrack employs a matching process based on motion and appearance similarity. Additionally, Kalman filters are utilized to predict object locations in subsequent frames, further enhancing tracking stability \cite{zhang2022bytetrack}. By effectively associating and refining detections, ByteTrack offers a robust tracking solution that complements fast and accurate object detectors like YOLOv8.
\subsection{Multi-View Approach}
Multi-view frameworks/approaches to MOT are not very common, the current most state-of-the-art approach was chosen and is described in this section. 

\subsubsection{DETR3D}
This research focused on accurate 3D object detection using information from multiple cameras, a task crucial for autonomous vehicles navigating complex environments \cite{wang2022detr3d}.  At its core, the method employs a novel 3D-to-2D query mechanism to directly link 3D positions to image features. This facilitates the detection of even partially occluded objects, as information is aggregated across views. It was innovative as it included a Transformer-based architecture designed to efficiently process multi-view data, leading to improved detection performance and reduced reliance on error-prone post-processing steps \cite{wang2022detr3d}.
\subsection{Findings and Research Gap}
The existing research underscores the potential of multi-view approaches for enhanced robustness in MOT, especially in addressing challenges like occlusions and variations in object appearance. Recent advancements in single-view trackers like FairMOT, and YOLOv8 demonstrate the power of balancing accuracy between detection and association, and refining label assignment strategies. However, a notable gap exists in the application of these techniques to animal tracking and specifically the complexities of fish behavior analysis in underwater environments. Existing multi-view methods such as DETR3D are also not designed to work with the outputs of these single-view trackers. This thesis thereby aims to fill in these gaps and help advance the field of not only MOT on animal data but also the field of multi-view MOT frameworks and methods. 

\section{Methodology}
\label{sec:methodology}
\par When working with FairMot and YOLOv8 it became clear that FairMot was unusable on the hardware and software available. FairMOT had an issue as the versions of Python and other requirements were outdated when compared to the versions running on local systems, Google Collab, and the UvA Snelius. Downgrading them and getting them to work was not possible, an essential part of this was also DCNv2 which was necessary to get FairMOT working and its requirements were also not in line with those of FairMOT itself or those of the systems used. YOLOv8 was the only model to work out of the box, it required the installation of some packages but worked almost immediately afterward. This led to YOLOv8 being chosen as the detection model for the research and multi-view framework. YOLOv8 was combined with ByteTrack as its tracker as ByteTrack had better results when compared to its native tracker BotSort. DETR3D was looked into but not able to be properly used with our stereo data coming from single-view models but was used as an inspiration for the proposed framework. 

\par Since the main goal and objective of our research is to research whether it is possible to adapt single-view model outputs in a way that can create a multi-view framework to improve tracking accuracy on fish data a framework was developed with a pipeline for using stereo videos and single-view multiple object detection and tracking models to create a new 3d view. 
\subsection{Experimental Setup}
\subsubsection{Dataset}
\par The dataset consists of three parts, the videos, the ground truth MOT files, and the stereo camera parameters. The video part contains 182 videos of which 91 are stereo videos, meaning there are two different camera views noted with \_1 and \_2. In total, there are  84,464 frames and 3,796 tracks. The short videos are 1080 x 1920 resolution, 240HZ, and contain 260 frames. The long videos are 1080 x 1920 resolution, 240HZ and 3117 frames for videos (129\_1, 161, 183, 231, 261\_1, 349, 406\_2) and 3118 frames for videos (129\_2, 349\_1, 406\_1). The short videos have been labeled by students and the long videos by the supervisors. 
\par The ground truth MOT file part contains .txt files that correspond to each video from the fish videos dataset part. These .txt files contain the tracking data per frame for all identified subjects. 
\par Each MOT file contains the following data: frame, id, top left x coordinate, top left y coordinate, bounding box width, and bounding box height. \cite{dendorfer2020mot20} Not all video pairs contain an overall equal amount of identified fish entities per video. This is because in certain frames fish entities may be visible in one of the two videos but not in the other one. In some cases, this is the case for the entire video leading to a mismatch in the amount of identified fish entities. It is also important to note that the fish entity IDs per video in a video pair are not equal, meaning that the fish with ID 1 in the \_1 video is not the fish with ID 1 in the \_2 video. 
\par The final part of the dataset consists of .mat files which contain the stereo camera parameters. This includes; distortion coefficients of camera 1 ('distortionCoefficients1'), distortion coefficients of camera 2 ('distortionCoefficients2'), the intrinsic matrix of camera 1 ('intrinsicMatrix1'), the intrinsic matrix of camera 2 ('intrinsicMatrix2'), the rotation of camera 2 with respect to camera 1 ('rotationOfCamera2'), and the translation of camera 2 with respect to camera 1 ('translationOfCamera2').
\subsubsection{The Multi-View Framework Overview}
\par The multi-view framework enhances single-view MOT by strategically integrating outputs from multiple cameras observing the same scene. With cameras positioned at slightly different angles or locations, while roughly at the same distance to the subjects, this system allows the creation of a new 3d coordinate system and unlocks new depth information compared to independent single-camera tracking. The following pipelines below demonstrate the processes with a short explanation for both the object detection training pipeline and the model detection and tracking execution pipeline. The entire process consists of cleaning and reformatting the raw data using Python scripts, training the model using YOLOv8, deploying the model to detect the fish entities using YOLOv8 and tracking them using ByteTrack, and finally using Python scripts for all post-track processes. The actual code files can be found in the GitHub link with extended comments further detailing how to run the code and how the code works. See Appendix E for the model training and fish detection and tracking pipeline visualizations.
\subsubsection{YOLOv8 Detection Model Training Pipeline}
\begin{enumerate}
\item \textit{Data Cleaning:}
\newline
The raw ground truth data comes in .txt files using the MOT format. These files will be referenced to as ground truth track files These files have multiple columns but no column headers and also contain a couple of columns with unnecessary data. This makes it less readable and clear so cleaning up the data is necessary before using it. This is done by adding column names and removing the last two columns that contain unnecessary data. The top left x coordinate and top left y coordinate are converted into center x and center y coordinates using the bounding box width and bounding box height. The bounding box width and height are converted into x offset and y offset. \par The new cleaned-up version is then saved with the "\_clean.txt" extension to its original name, see Appendix B. 
\begin{equation}
x_{\text{center}} = x_{\text{tl}} + \frac{w}{2}
\end{equation}

\begin{equation}
y_{\text{center}} = y_{\text{tl}} + \frac{h}{2}
\end{equation}
\begin{equation}
x_{\text{offset}} = \frac{w}{2}
\end{equation}

\begin{equation}
y_{\text{offset}} = \frac{h}{2}
\end{equation}
where:

\begin{itemize}
    \item $ x_{\text{tl}} $: x-coordinate of the top-left corner.
    \item $ y_{\text{tl}} $: y-coordinate of the top-left corner.
    \item w: width of the bounding box.
    \item h: height of the bounding box.
\end{itemize}

\item \textit{Data Formatting:}
\newline
The YOLOv8 model requires training and validation data in order to be trained. YOLOv8 needs the videos to be split up into separate images per frame with a corresponding .txt file per frame containing the class\_label, center\_x, center\_y, width, and height. The center\_x and center\_y are equal to the x and y from the cleaned-up ground truth track files. The width is calculated using, width = 2 * x\_offset and the height is calculated with, width = 2 * y\_offset. The frame and cleaned-up ground truth track file are saved as "frame\_1.jpg" and "frame\_1.txt" with each pair having its own unique number ID to show that they are a match. The video frames are then saved in an images folder and the new cleaned-up ground truth track files are in a labels folder as they are seen as labels by YOLOv8. This has to be done for both the training and validation data. The folders then need to be restructured manually to conform with the YOLOv8 requirements to be read correctly. A dataset folder is created that contains images and labels folders, both of which contain a train and val folder. Lastly, a Data file is created manually which needs to contain the following code seen below and saved as data.yaml.
\lstset{
  language=Python,
  basicstyle=\scriptsize\ttfamily, 
  numbers=left,
  numberstyle=\tiny,
  commentstyle=\color{gray},
  backgroundcolor=\color{white} 
}
\begin{lstlisting}[caption={data.yaml File}]
train: images/train  # train images (relative to 'path') 
val: images/val  # val images (relative to 'path') 
nc: 1  # Number of classes (just fish)
names: ['fish']  # Your class names
\end{lstlisting}
\item \textit{YOLOv8 Model Training and Exporting}
\newline
To Train the model the following line of code is used to train the model in the most basic form: "model.train(data="data.yaml", epochs=100, imgsz=1920)" \cite{ultralyticsHome}. The data.yaml file contains the instructions as to where the labels and images are, the epochs denote how many training epochs are allowed and imgzs determine the image size. Other arguments can be passed to the model training and can be found with in-depth descriptions in the official YOLOv8 documentation. The code automatically creates a runs folder locally and saves the run with in it the best-performing model. 
\newline
\par YOLOv8 training revolves around feeding the model with labeled image data and adjusting its internal parameters to minimize the difference between the model's predictions and the actual labels. This iterative process refines the model's ability to identify objects in unseen images. \cite{ultralyticsHome,lou2023dc,github_akashAD98_yolov8,github_ultralytics_2789,github_ultralytics_4684}
\newline
\par The optimization of YOLOv8 is guided by its loss function. The loss function used by YOLOv8 is a combination of several terms:
\begin{itemize}
    \item Bounding Box Loss (CIoU Loss): This term measures the distance between the predicted bounding box and the ground truth box. It penalizes for both size and location errors, using the recently introduced CIoU (Complete Intersection over Union) loss for accuracy. \cite{ultralyticsHome,lou2023dc,github_akashAD98_yolov8,github_ultralytics_2789,github_ultralytics_4684}
\begin{equation}
\small
\text{CIOU Loss} = \text{IoU} + \lambda_{txt} (v - \text{IoU}^{\alpha}) + \alpha \cdot v \cdot (1 - \text{IoU}^{\alpha}) \left( 4 \arctan \left(\frac{tx - tw}{ty - th} \right) \right)^2
\end{equation}    
Where:
\begin{itemize}
\item IoU: Intersection over Union between the predicted and ground truth boxes
\item $\lambda_{txt}$ :Balancing parameter for the excess area term (v)
\item $\alpha$: Balancing parameter for the aspect ratio term
\item v: \begin{equation}
\small 
v = \frac{(w^g - w) \cdot (h^g - h)}{(w^g \cdot h^g + \varepsilon)}
\end{equation}
\begin{itemize}
    \item w,h: Width and height of the predicted bounding box
    \item wg,hg: Width and height of the ground truth bounding box
    \item $\varepsilon$: Small constant to avoid division by zero
\end{itemize}
\item tx, ty: Center coordinates of the predicted bounding box
\item tw, th: Center coordinates of the ground truth bounding box
\end{itemize}
    \item Classification Loss (Cross-Entropy Loss): This term penalizes the model for incorrectly classifying the object within the bounding box. It utilizes cross-entropy loss, a common choice for classification tasks. \cite{ultralyticsHome}
    \begin{equation}
    \small
        \text{Classification Loss} = - \sum_{i} (y_i \cdot \log(p(i)))
    \end{equation}
    Where:
    \begin{itemize}
        \item $\sum_i$: Summation over all classes
        \item $y_i$: Binary indicator (1 for ground truth class, 0 for others)
        \item p(i): Predicted probability for class i
    \end{itemize}
    \item Objectness Loss (Binary Cross-Entropy): This term distinguishes between locations with objects (foreground) and those without (background). It employs binary cross-entropy loss for this binary classification task. \cite{ultralyticsHome}
    \begin{equation}
    \small
        \text{Objectness Loss} = - \sum [(1 - c) \cdot \log(1 - p(obj)) + c \cdot \log(p(obj))]
    \end{equation}
    where:
    \begin{itemize}
        \item $ \sum$: Summation over all cells in the image grid
        \item c: Ground truth object confidence score (1 for containing an object, 0 for background)
        \item p(obj): Predicted probability of a cell containing an object
    \end{itemize}
\end{itemize}
The total loss is calculated by summing these individual losses, providing a comprehensive error signal for the model to learn from during backpropagation.
\newline
\par Due to low performances when training and optimizing the model and having limited computing resources a model trained by Dr Fatemeh Karimi Nejadasl was chosen to be used for further use instead of any of the locally trained models. 
\end{enumerate}
\subsubsection{Fish Detection and Tracking Pipeline}
\begin{enumerate}
    \item \textit{Single-View Tracking Using YOLOv8 Combined with ByteTrack:}
\newline 
\par The Video inputs are fed into the YOLOv8 model one by one and the ByteTrack tracker is selected for the tracking part of the model deployment. By default, the output is a video containing the tracking data overlayed and a results variable which in it contains bounding boxes information, class label information, the confidence score, and the track ID. To extract this and convert it into a workable DataFrame similar to the original ground truth track files data a Python script is used to create a dictionary from the results data which can then be converted to a pandas DataFrame and saved as a CSV file. The bounding boxes contain xmin, ymax, ymin and ymax coordinates but since we want center-x, center-y, x-offset and y-offset this also needs to be converted calculated, and matched to the correct ID and Frame. The xmin, ymin, xmax and ymax are taken from the bounding boxes, and the new values needed are calculated in the following ways:
\begin{itemize}
    \item center-x = (xmin + xmax) / 2
    \item center-y =  (ymin + ymax) / 2
    \item x\_offset = |center-x - xmin|
    \item y\_offset = |center-y - ymin|
\end{itemize}

\item \textit{Post Track Fish ID Re-Identification:}
\par When a fish entity has not been identified for a couple of frames and then is identified again it is often seen as a new fish in the output. This means that the output has multiple fish IDs which should actually belong to the same fish. To combat this a post-track re-identification part is added to the pipeline in the form of a Python script that is not part of the native ByteTrack implementation. This tries to re-identify fish that have been assigned new IDs and resets their new ID back to their original ID. The post-track re-identification (post-track-re-id) part also aims to remove as much false positives as possible by closely examining the amount of frames for which a fish entity is identified. The first step in the post-track-re-id process is determining which fish entities are not present for all frames in a video. These fish entities are then deemed as possible candidates for post-track-re-id. The next step is locating fish entities that "disappear" and have a "new" fish (with x and y coordinates) entity "appear" in close proximity to them within a 100-frame window. Close proximity is when a "new" fish entity's distance is within 50 pixels from the "old" fish entity (with prev\_x and prev\_y as its x and y coordinates. This distance is calculated using 
\begin{equation}
\small
    distance = ((x - prev_x) ** 2 + (y - prev_y) ** 2) ** 0.5
\end{equation}
\par Then it is checked if this "new" fish entity was present at the same time as the "old" fish entity. If the "new" fish was present at the same time as the "old" fish entity for 10 or fewer frames and these 10 or fewer frames are the last 10 frames in which the "old" fish entity was present then it can be assumed that the "new" fish entity is in fact the "old" fish entity but with a new ID because it was not identified and tracked in the in-between frames. The "new" fish entity then gets its ID reset to the "old" fish entity ID. This whole process is done within a loop until there are no more cases for which this is true. When the loop has finished any fish entities that appear in less than 30 frames are removed as they are assumed to be false positives. Finally, all IDs are reset to the lowest possible numbers to ensure there are no large numerical discrepancies between the fish IDs after which the new DataFrame is saved as a CSV file containing the changes made by the post-track-re-id process. 
\item \textit{Fish Tracking Overlay Video Output:}
\par Using the tracks saved and the original video, in a Python script the bounding boxes are calculated and overlayed on the original video together with the corresponding fish entity IDs and saved as new videos. 

\item \textit{Post-Track Calibration-Based Matching:}
\par To be able to match the fish entity IDs from the video pairs stereo matching using epipolar geometry is performed. This process is built out of 4 sub-processes:
\begin{itemize}
    \item Stereo Camera Calibration and Feature Matching: 
    \newline
We organize image pairs and iterate through filenames. For each pair, we match images based on frame number, then retrieve camera data for distortion correction. We prepare images and identify features like corners. These features are refined based on camera data. We then find matching features between images and filter out unreliable ones. Finally, we calculate the fundamental matrix, a mathematical relationship between the cameras based on good matches. This matrix, along with good features, is stored for each image pair.\cite{yamashita2003camera,stankiewicz2018multiview,diaz2022stereo,hartley2003multiple,hirschmuller2007stereo}
\item Epipolar Line Prediction and Matching:
\newline
The fundamental matrix predicts where a left keypoint's match lies in the right image. We calculate an epipolar line on the right image based on this prediction. Keypoints near this line are considered potential matches.
\cite{yamashita2003camera,stankiewicz2018multiview,diaz2022stereo,hartley2003multiple,hirschmuller2007stereo}
\item Geometric Matching Refinement:
\newline
To refine the matching process and identify the most likely corresponding point, we exploit the geometric relationship between the cameras encoded in the fundamental matrix. We calculate the distance between each potential match in the right image and the predicted epipolar line.
A threshold is applied to filter out points with significant deviations from the epipolar line, potentially caused by noise or mismatches.
\cite{yamashita2003camera,stankiewicz2018multiview,diaz2022stereo,hartley2003multiple,hirschmuller2007stereo}
\item Correspondence Collection and Evaluation:
\newline
The process iterates through all keypoints in the left image for each frame. For each left keypoint, a corresponding point with the minimal distance to the epipolar line within the search area is identified (if one exists). This point is considered the most likely match in the right image.
The left and right keypoint IDs, along with the difference between the match and the epipolar line (representing the matching error), and the frame number are stored in a DataFrame and saved as an CSV file.
\cite{yamashita2003camera,stankiewicz2018multiview,diaz2022stereo,hartley2003multiple,hirschmuller2007stereo}
\end{itemize}
\item \textit{Post-Track Frequency Match Pair Determination:}
\newline
The output from the previous part contains the best match per frame for each fish entity. To find out which of those is the True best match we need to determine which match is the most occurring per fish entity and set this as the True match for that given fish entity. This is then saved in a DataFrame such as seen in appendix A.

\item \textit{Post-Track 3D Coordinate Creation:}
\newline
To be able to create 3D coordinates the x and y coordinates are necessary per frame for each fish entity in both videos together with the camera parameters. To do this the separate CSV files for the video pairs are loaded and merged using the StereoMatching DataFrame from before to map the IDs to each other. We then iterate through the set of unique frame numbers present in the left image DataFrame (video \_1 of the pair). For each frame, we retrieve a temporary DataFrame containing information about the keypoints in both the left and right images for that specific frame. From here we iterate through each ID present in the left image DataFrame for the current frame and retrieve the matching IDs from the right image DataFrame. If the match exists we extract the x and y coordinates for both keypoints. Then the camera calibration parameters are leveraged to perform image triangulation to estimate the 3D location corresponding world point. See Appendix D for a visual illustration.
\cite{yamashita2003camera,stankiewicz2018multiview,diaz2022stereo,hartley2003multiple,hirschmuller2007stereo}

The homogeneous coordinates obtained from the triangulation are then converted to Cartesian coordinates and the x,y, and z coordinates of the reconstructed 3D point are extracted and saved in a new DataFrame. This is then done for each Frame in video 1 resulting in a DataFrame containing new x,y and z coordinates for each fish entity that is present in both videos for the given frame for each frame. This new DataFrame is then saved as a CSV file to be used in the evaluation and data exploration.
\item \textit{Final Data Visualization}
\newline
To be able to properly analyze the data generated in step 6 it is necessary to generate the following graphs and statistics: fish trajectories in 3d, fish speed over time, fish acceleration over time, fish path length, fish spatial distribution, fish density, temporal patterns over time, and fish depth over time. Example output can be seen in appendix C. These are used to research the fish entity movement patterns and possibly derive behavioral patterns and compare and evaluate what new insights can be made by using the 3D coordinates when compared to the single-view 2D coordinates. 
\end{enumerate}
\subsection{Evaluation}
\par We evaluated fish detection performance using precision, recall, and bounding box size \cite{powers2020evaluation}, the percentage of fish identified within the pre-defined margins, the amount of fish identified but not within the pre-defined margins, the amount of fish not identified and the number of false positives. A detection was considered a correct identification within the margin if the center of its bounding box fell within half the average bounding box size of the ground truth fish. We used a larger margin, twice the average size, to capture near misses where the detection was close but not perfectly centered. This two-tier approach helps differentiate between precise detections and those that are close but slightly off-center. Unmatched detections (no corresponding ground truth) and missed fish (no detection at all) were counted as errors. Additionally, we employed the comprehensive evaluation metrics from the FairMOT paper (HOTA, DetA, AssA, MOTA, IDF1) for a more in-depth analysis. \cite{gao2023memotr,yu2023motrv3,Luiten_2020,bernardin2008evaluating,ristani2016performance}. 
\par To be able to correctly map the fish IDs from the tracking results to the ground truth an ID mapping function was made that had 100\% accuracy in mapping the tracking IDs to the ground truth IDs. 
\par The accuracy of the stereo matching component was also evaluated as this played a major part in the creation of the 3D coordinates. This was done by comparing the fish ID mapping from the stereo matching to the ground truth mapping which was done by hand.
\par The number of fish entities identified before the re-identification step and after the re-identification step was also compared to the ground truth to evaluate the performance of the re-identification part of the pipeline.
\par Finally, for each video for each fish, the frames for which that fish was not identified were calculated and saved into a JSON dictionary. This was done to help visualize when fish that were previously identified were not identified and for how many frames. To achieve this we compared the set of all frames per fish ID from the tracking model with the ground truth data and determined the difference in frames between the two.
\subsubsection{HOTA}
\begin{equation}
\small
\text{HOTA} = \left( \prod_{k=1}^{K} \left( \text{DA} \times \text{AA} \times \text{LA} \right)^{\frac{1}{K}} \right)^{\frac{1}{3}}
\end{equation}

where:

\begin{itemize}
  \item $\text{DA}$ (Detection Accuracy) measures how well the tracker detects objects, and is defined as:
  \begin{equation}
  \small
  \text{DA} = \frac{\text{TP}}{\text{TP} + \text{FN} + \text{FP}}
  \end{equation}
  \item $\text{AA}$ (Association Accuracy) measures how well the tracker associates detections over time, and is defined as:
  \begin{equation}
  \small
  \text{AA} = \frac{\text{Correctly Associated Pairs}}{\text{Total True Positives}}
  \end{equation}
  \item $\text{LA}$ (Localization Accuracy) measures how accurately the tracker localizes the detected objects, and is defined as:
  \begin{equation}
  \small
  \text{LA} = \frac{\sum_{i=1}^{N} \left(1 - \frac{\text{Localization Error of } i}{\text{Total Possible Localization Error}}\right)}{N}
  \end{equation}
\end{itemize}

\cite{Luiten_2020}
\subsubsection{DetA}
\begin{equation}
\small
\text{DetA} = \frac{\text{TP}}{\text{TP} + \text{FN} + \text{FP}}
\end{equation}
\cite{Luiten_2020}
\subsubsection{AssA}
\begin{equation}
\small
\text{AssA} = \frac{\text{Correctly Associated Pairs}}{\text{Total Number of Associations}}
\end{equation}
where:
\begin{itemize}
  \item $\text{Correctly Associated Pairs (CAP)}$ are pairs of detections that are correctly identified as the same object across consecutive frames.
  \item $\text{Total Number of Associations (TNA)}$ is the total number of associations that the tracking algorithm makes, including both correct and incorrect associations.
\end{itemize}

\cite{Luiten_2020}
\subsubsection{MOTA}
\begin{equation}
\small
\text{MOTA} = 1 - \frac{\sum_t (\text{FN}_t + \text{FP}_t + \text{IDSW}_t)}{\sum_t \text{GT}_t}
\end{equation}

where:
\begin{itemize}
  \item $\text{FN}_t$ is the number of false negatives (missed detections) at time $t$.
  \item $\text{FP}_t$ is the number of false positives (incorrect detections) at time $t$.
  \item $\text{IDSW}_t$ is the number of identity switches at time $t$.
  \item $\text{GT}_t$ is the number of ground truth objects at time $t$.
\end{itemize}
\cite{bernardin2008evaluating}
\subsubsection{IDF1}
\begin{equation}
\small
\text{IDF1} = 2 \times \frac{\text{Precision} \times \text{Recall}}{\text{Precision} + \text{Recall}}
\end{equation}
\cite{ristani2016performance}

\section{Results}
\label{sec:results}
\subsection{Single-View Detection and Tracking MOT Metric Results}
\begin{table}[H]
    \centering
    \caption{MOT Evaluation Metrics: Full table in Appendix F}
    \small 
    \setlength{\tabcolsep}{4pt} 
    \begin{tabular}{l l l l l l l l}
    \hline
        ID & Precision & Recall & HOTA& MOTA & AssA& DetA& IDF1 \\ \hline
        8\_1\_tr.csv & 1.0 & 0.386 & 0.076 & 0.385 & 1.0 & 0.007 & 0.556 \\ 
        8\_2\_tr.csv & 1.0 & 0.43 & 0.08 & 0.429 & 1.0 & 0.013 & 0.6 \\ 
        10\_1\_tr.csv & 0.957 & 0.363 & 0.076 & 0.379 & 1.0 & 0.022 & 0.55 \\ 
        10\_2\_tr.csv & 0.799 & 0.402 & 0.082 & 0.454 & 1.0 & 0.002 & 0.668 \\ 
        13\_1\_tr.csv & 1.0 & 0.333 & 0.07 & 0.315 & 1.0 & 0.06 & 0.498 \\
        Average* & 0.868 & 0.480 & 0.100 & 0.506 & 0.994 & 0.023 & 0.677 \\
 \hline
    \end{tabular}
\end{table}
\begin{figure}[H]
    \centering
    \includegraphics[width=4cm]{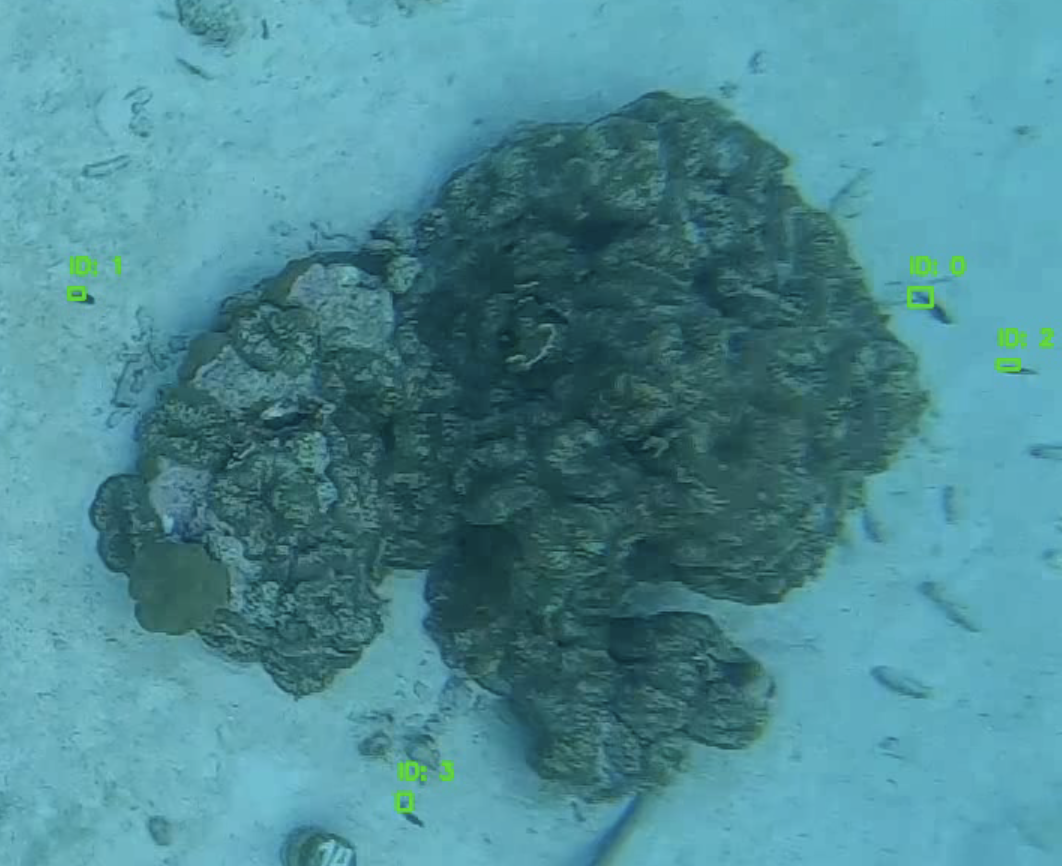}
    \includegraphics[width=4cm]{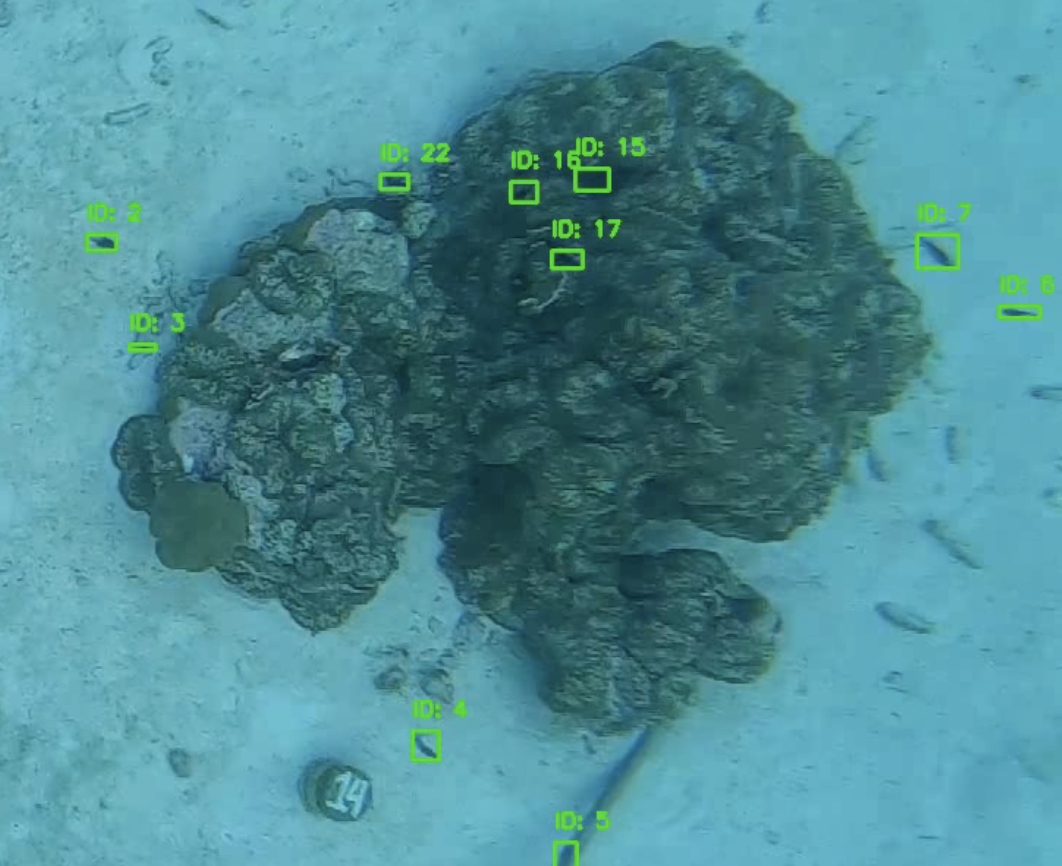}
    \caption{Model Output VS Ground Truth Identification for a Random Frame of Video 23\_1}
    \label{fig:ReID}
\end{figure}
\par We use pre-established MOT evaluation metrics to evaluate the performance of the model on each video track to determine the overall performance. As seen in Table 1 and Appendix F, the YOLOv8 fish tracking model achieved a precision of 0.868, indicating good accuracy in identified fish. However, the low recall of 0.480 reveals a significant limitation, as the model misses nearly half of the actual fish. This is further corroborated by the low HOTA (0.100) and MOTA (0.506) scores, highlighting issues with overall tracking performance. While the high AssA (0.994) suggests consistency in tracking detected fish, the very low DetA (0.023) signifies a major challenge with inaccurate fish localization. This is highlighted by Figure 1, visualizing the output for a randomly selected frame in video 23\_1.
\begin{table}[H]
    \centering
    \caption{YOLOv8 Data Set Performance Comparison}
    \small 
    \setlength{\tabcolsep}{4pt} 
    \begin{tabular}{l l l l l l l l}
    \hline
        Data Set & HOTA& MOTA & AssA& DetA& IDF1 \\ \hline
        MOT17   & 0.631 & 0.803 & 0.620 & 0.645 & 0.773 \\
        Fish Data & 0.100 & 0.506 & 0.994 & 0.023 & 0.677 \\\hline
    \cite{zhang2022bytetrack,xiao2023multi}
    \end{tabular}
\end{table}
\par Table 2 compares the performance of our YOLOv8 fish tracking model with results on another commonly used dataset in MOT evaluations, MOT17. While our model achieves a high AssA, indicating consistency in tracking detected fish, it falls short in other metrics compared to the other dataset. Notably, the Fish Data set has the lowest HOTA and MOTA scores, suggesting a significant struggle with overall tracking accuracy. This is further corroborated by the very low DetA, signifying a major challenge with accurate fish localization. In comparison, MOT17 shows considerably better performance in both tracking accuracy (HOTA and MOTA) and detection accuracy (DetA). This suggests that YOLOv8 models generalize well to pedestrian or generic object-tracking tasks but require further optimization for the specific challenges of fish tracking in underwater environments even when trained on these specific underwater environments.

\subsection{Single-View Detection and Tracking Specific Evaluation Results}
\begin{table}[H]
    \centering
    \caption{Margins Evaluation Metrics: Full table in Appendix F}
    \small 
    \setlength{\tabcolsep}{4pt} 
    \begin{tabular}{l l l l l}
    \hline
        ID & Within Margin & Not Within Margin & Not Identified & False Positive \\ \hline
        8\_1\_tr.csv & 0.385 & 0.0 & 0.611 & 0.0 \\ 
        8\_2\_tr.csv & 0.428 & 0.0 & 0.568 & 0.0 \\ 
        10\_1\_tr.csv & 0.361 & 0.002 & 0.633 & 0.016 \\ 
        10\_2\_tr.csv & 0.401 & 0.0 & 0.595 & 0.101 \\ 
        13\_1\_tr.csv & 0.332 & 0.0 & 0.664 & 0.0 \\ 
        Average* & 0.478 & 0.015 & 0.500 & 0.046 \\\hline
    \end{tabular}
\end{table}
\par In Table 3 we see a subset of how much percent of the fish entities are identified within the pre-defined margins, not within the pre-defined margins, not identified at all, and the percentages of false positives. On average less than half of the fish entities are detected within the margins and half are not identified at all. This shows us that the model does not manage to detect all fish entities and often only sees part of the fish entities as actual fish. This can be seen in Figure 2 where on the left we have the YOLOv8 model and on the right, we have the ground truth version. The model does not capture the entire fish entity as the ground truth does. This leads to there being a center x and y that is offset rather much from the ground truth x and y. In Figure 3 we can also see how the model often misses fish entities that should be seen. Here we see a fish entity with ID 4 that is marked in the ground truth but not detected in our model. Lastly, the amount of false positives is low but there are still instances of objects being identified as fish when they should not be which can be seen in Figure 4.

\begin{figure}[H]
    \centering
    \includegraphics[width=3cm, height=3cm]{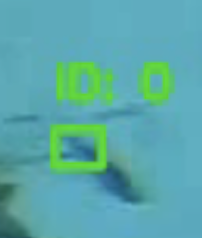}
    \includegraphics[width=3cm, height=3cm]{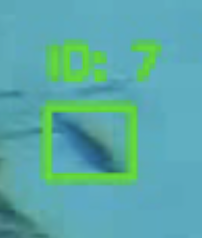}
    \caption{Fish margins of fish entity 7 in Video 23\_1: The left image shows the YOLOv8 output and the right image shows the ground truth.}
    \label{fig:ReID}
\end{figure}
\begin{figure}[H]
    \centering
    \includegraphics[width=3cm, height=3cm]{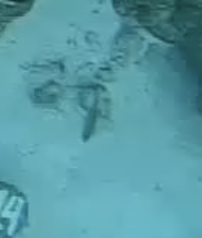}
    \includegraphics[width=3cm, height=3cm]{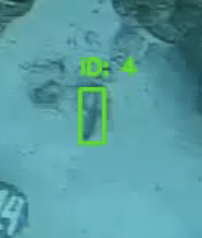}
    \caption{Missed Identification of fish entity 4 in Video 23\_1: The left image shows the YOLOv8 output and the right image shows the ground truth.}
    \label{fig:ReID}
\end{figure}
\begin{figure}[H]
    \centering
    \includegraphics[width=3cm, height=3cm]{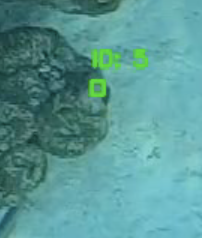}
    \includegraphics[width=3cm, height=3cm]{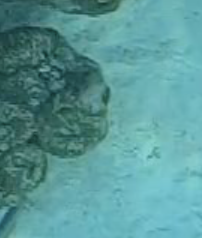}
    \caption{False Positive Identification of fish entity in Video 10\_2: The left image shows the YOLOv8 output and the right image shows the ground truth.}
    \label{fig:ReID}
\end{figure}
\subsection{Re-Identification code Evaluation Results}

\begin{figure}[H]
    \centering
    \includegraphics[width=4cm]{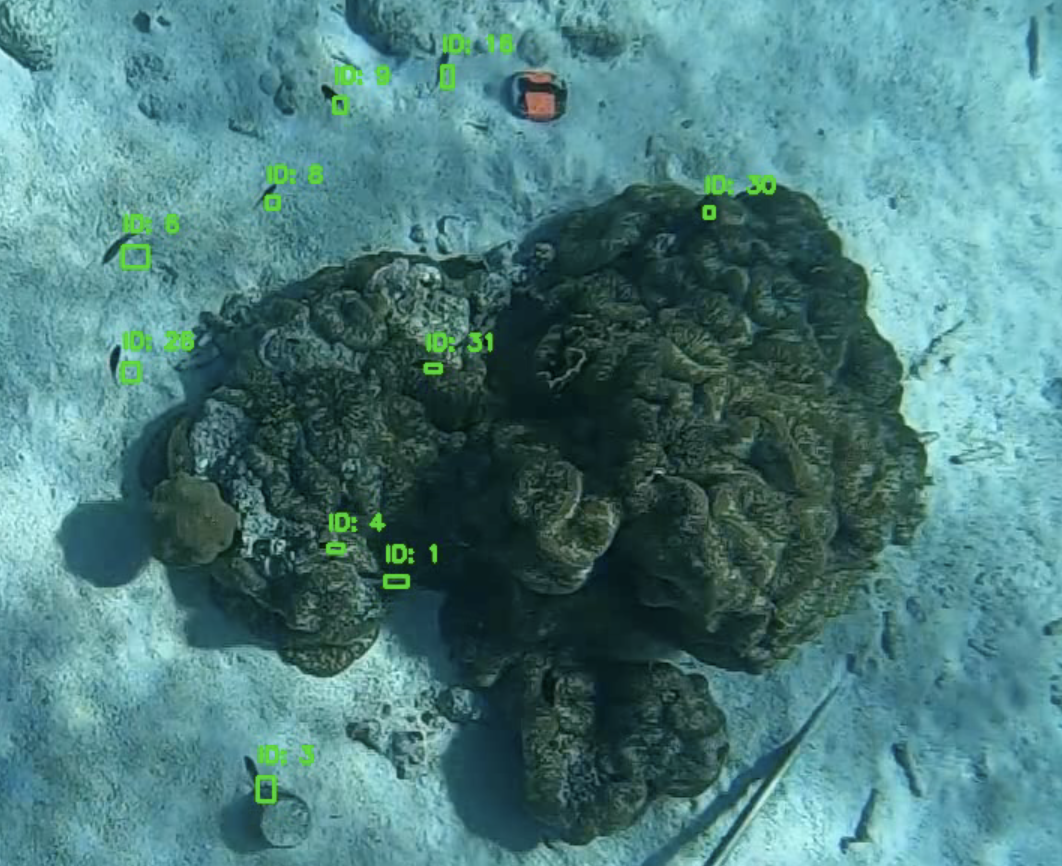}
    \Description{Tracking result before re-identification for video 129_1.}
    \includegraphics[width=4cm]{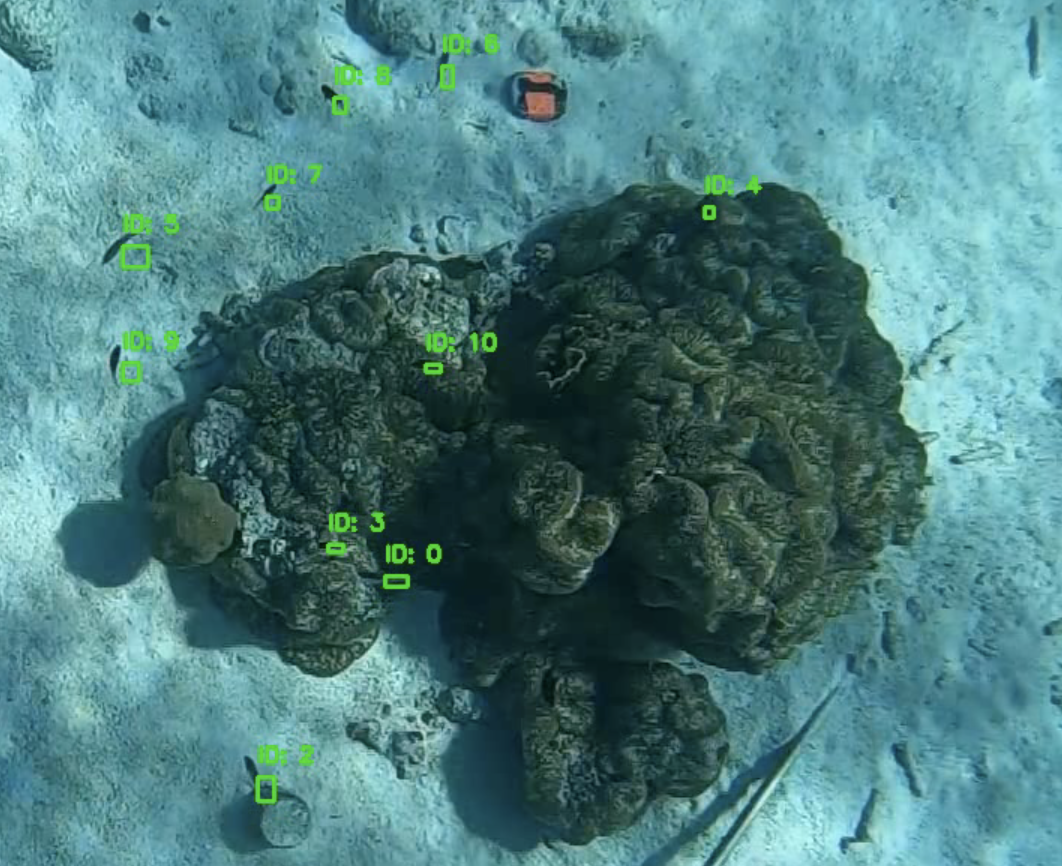}
    \caption{Tracking Before and After Re-Identification Video 129\_1}
    \label{fig:ReID}
    \hspace*{-2.5cm} 
\end{figure}
\begin{table}[H]
    \centering
    \caption{ID Count Evaluation: Full table in Appendix F}
    \small 
    \setlength{\tabcolsep}{4pt} 
    \begin{tabular}{l l l l}
    \hline
        ID & GT IDs & Before Re-ID & After Re-ID \\ \hline
        8\_1\_tr.csv & 9 & 12 & 6 \\ 
        8\_2\_tr.csv & 9 & 11 & 5 \\ 
        10\_1\_tr.csv & 9 & 16 & 7 \\ 
        10\_2\_tr.csv & 9 & 23 & 13 \\ 
        13\_1\_tr.csv & 13 & 17 & 6 \\ \hline
    \end{tabular}
\end{table}

Initially, the model did not do a satisfactory job of tracking the fish amongst all frames leading to many fish entities having multiple IDs. To combat this we created a re-identification process which takes place after the model has created its output. This led to a drop in the amount of IDs in the output with performances varying per video. Table 4 shows a subset of the amount of IDs that should be present versus the initial amount created by the model and the amount after applying the re-identification process. On average when looking at the full table in Appendix F the amount of Fish IDs went down by 43.3\% after having applied the re-identification.
\par The numbers on their own however do not paint the full picture as for example for video 13\_1 the amount before the re-identification is 17 which is more than the ground truth amount and after the re-identification drops down to 6. This does not suggest that fish entities have no longer been tracked but rather that many subjects that seemed to be unique were in fact the same subject. The re-identification process helps in better visualize how many fish entities are not identified at all and helps us in creating better tracking results. This can also be seen in Figure 5 where the IDs numerical values in our models output are way higher than in the ground truth suggesting more fish IDs then there are in fact due to fish not being tracked along all frames. 

\subsection{3D Coordination Creation Results}
\begin{table}[H]
    \centering
    \caption{3D Coordinates Table Snippet Video 129} 
    \small 
    \setlength{\tabcolsep}{4pt} 
    \begin{tabular}{l l l l l l}
    \hline
        frame & left\_id & right\_id & x & y & z \\ \hline
        1 & 0 & 5 & 475.641 & -251.609 & 2175.999 \\ 
        1 & 1 & 6 & 358.941 & -255.299 & 2170.809 \\ 
        1 & 2 & 3 & 1187.891 & -270.103 & 2587.299 \\ 
        1 & 3 & 0 & 273.150 & -478.047 & 2606.051 \\ 
        1 & 4 & 2 & 634.803 & -551.119 & 2809.724 \\ \hline
    \end{tabular}
\end{table}
\begin{figure}[h]
    \centering
    \includegraphics[width=5cm]{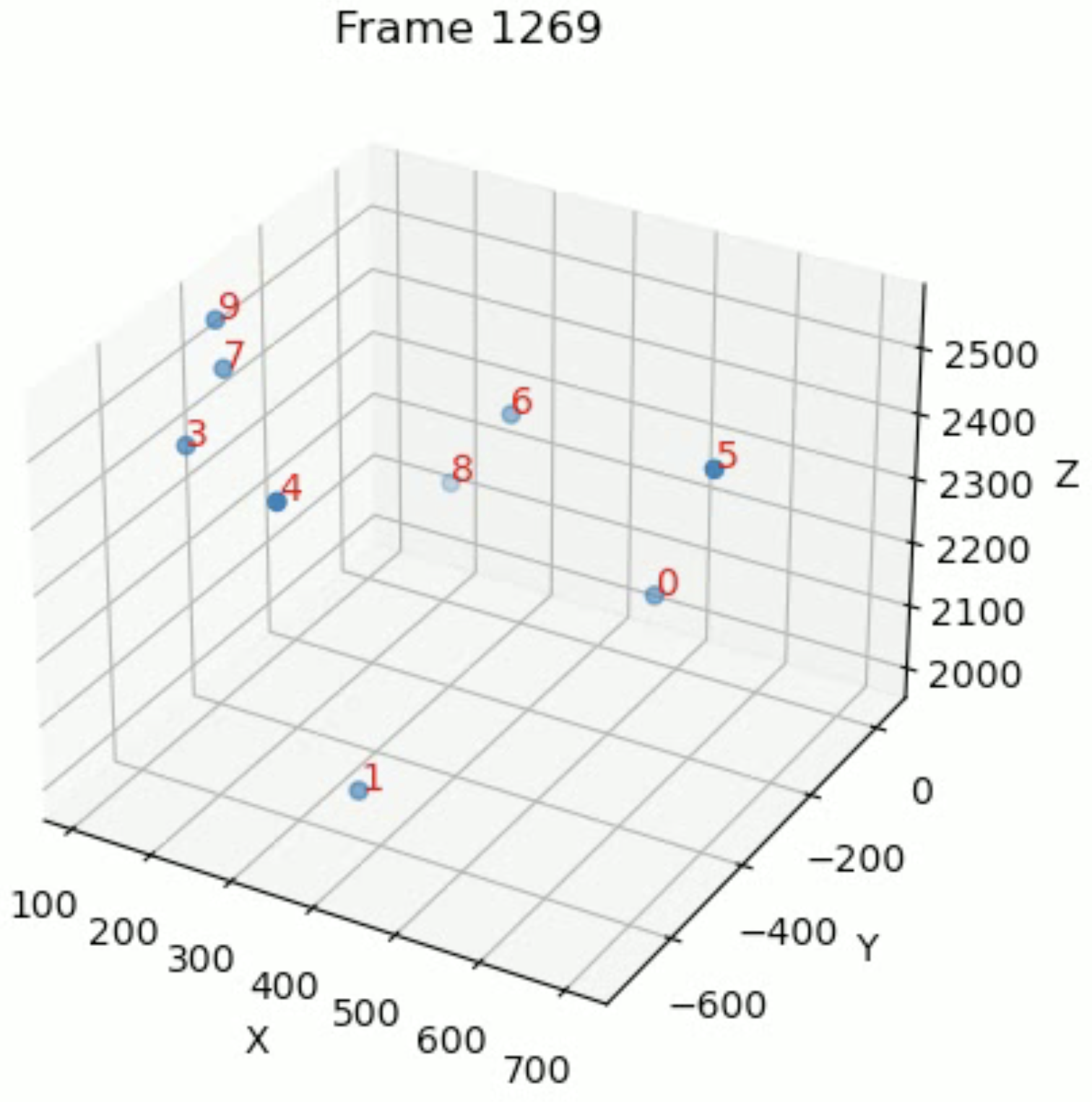}
    \caption{3D Coordinates Visualization of Video 129 For Frame 1269}
    \label{fig:ReID}
\end{figure}
\par Our model creates a new set of 3D coordinates for the fish entities to be able to add depth information which was missing from the single-view results. The single-view results only generate 2D information and therefore are missing depth information. The 3d coordinates however can only be made for frames in which both videos see the same fish entity and have been correctly stereo-matched. The output therefore is not complete for all frames as not all fish entities are seen in each frame in both videos. A subset of the output generated can be seen in Table 5. With this output we can then create a 3D graph helping visualize the fish movements between 3D space, Figure 6 shows this for a randomly selected Frame from Video 129.
\subsection{Stereo Matching Evaluation Results}
\begin{table}[H]
    \centering
    \caption{Subset of Stereo-Matching Accuracy} 
    \small 
    \setlength{\tabcolsep}{4pt} 
    \begin{tabular}{l l l l l l}
    \hline
        Video & Correctly Matched IDs & Video Length in Frames \\ \hline
        8 & 22\% & 258  \\ 
        23 & 36\% & 258  \\ 
        129 & 95\% & 3117  \\ 
        406 & 100\% & 3117 \\ \hline
    \end{tabular}
\end{table}
The stereo-matching part of our pipeline is essential in the matching and mapping of the IDs between the video pairs and is used for the creation of the 3D coordinates as they require pairs of coordinates from both video pairs. The stereo-matching did not end up being robust as it performed poorly in short video tracks and well in long video tracks as seen in Table 6. This meant that the matches for the short video tracks were unusable and had to be re-matched manually to be able to be used for the creation of 3D coordinates. The long tracks however did perform well enough to be able to be used without manual matching. 
\subsection{Fish Track Completeness Evaluation Results}
\begin{table}[H]
    \centering
    \caption{Amount of Missing Frames Snippet Video 8\_1} 
    \small 
    \setlength{\tabcolsep}{4pt} 
    \begin{tabular}{l l l l l l}
    \hline
        ID & Amount of Missing Frames \\ \hline
        2 & "Missing match"  \\ 
        3 & 222  \\ 
        4 & "Missing match"  \\ 
        5 & 134 \\ 
        6 & 56  \\ 
        7 & 1 \\ 
        8 & 16  \\ 
        17 & "Missing match"  \\ 
        18 & "Missing match"  \\ \hline
    \end{tabular}
\end{table}
\par To be able to further analyse the tracking of the fish entities we needed to evaluate how many frames they were detected in and tracked for when compared to the ground truth. This was written to a JSON file containing a dictionary containing the exact frames for which each fish entity was not tracked per video and to a JSON file containing a dictionary containing the amount of frames for which each fish entity was not tracked per video. A snippet of this can be seen in Table 7, "Missing Match" signifies that the fish was not tracked at all, 0 means it was tracked for all frames and the number values indicate how many frames a fish entity was not tracked in. 
\par On average 26.6\% of the ground truth fish entities are not detected at all in the tracking results, meaning that they fish were completely missed by out model. Only 18.9\% of the fish entities were fully tracked and detected for all frames in our model and 45.5\% of the fish entities were tracked but not for all of the frames they should have been tracked in. 

\section{Discussion}
\label{sec:discussion}
\par In this section, we will elaborate on how our chosen methodology
relates to existing works and how we interpret our results. Additionally, this section will be used to acknowledge any limitations in our experimental design to ensure a clear understanding of the study's scope and potential areas for further research.
\subsection{Reflection On Model}
Based on the results gathered from the evaluation the model on its own without the implementation of the re-identification process does not perform well. The model almost always fails to get the amount of fish correct due to fish entities not being correctly tracked and therefore getting multiple IDs. This is overcome however with the implementation of the re-identification process which improves its results. This was to be expected tho due to the rapid appearance changes of the fish when they rotate their body. This makes them appear different and often unidentifiable. The light refractions underwater also play a role in trying to correctly keep track of the fish together with the fact that fish often can swim under each other or close to each other making them no longer appear as singular fish and disturbing the tracking process. 

\par The actual identification of fish entities does not perform well when compared to the identification of the subjects in the MOT17 dataset. The model does not identify the fish entities within the margins nearly enough with the correct identifications being less than 50\%. This was expected as the fish are visually similar to their backgrounds and small with hardly any unique recognizable features for the model to learn. This combined with the fact that their appearance changes drastically when they change their rotation makes for bad identification performance \cite{Er_2023}. 
\par Finally the Stereo-matching was not robust enough and performed poorly on the small videos and well on the long videos making it only usable on longer videos. This was not expected as in other papers regarding stereo matching and epipolar geometry had rather good results \cite{yamashita2003camera,stankiewicz2018multiview,diaz2022stereo,hartley2003multiple,hirschmuller2007stereo}.  
\subsection{Limitations}
Our model was limited due to multiple factors, firstly the dataset was not large or diverse. Many of the videos were extremely similar making our model not generalized enough. The videos themselves were of low resolution making it harder to properly distinguish between fish entities and their background, the water conditions were also not always ideal in the videos, often dust particles moved around in ways similar to fish. Due to the footage being underwater, there was low color contrast, almost all objects in the videos had the same color and texture. Due to this, it would be advisable for further research to be done using other camera setups, experimenting with more diverse camera angles, sharper and more high-resolution footage, more diverse data, and implementing methods to increase the contrast between the colors.  

\section{Conclusion}
\label{sec:conclusion}
\par This research tries to leverage stereo videos to create 3D data on underwater fish entities using a single-view detection and tracking model. This research aimed to answer to what extent a proposed multi-view multiple-object tracking framework using single-view multi-object trackers can be effectively implemented and adapted to overcome the challenges of fish tracking in datasets characterized by small and visually similar fished with rapidly changing appearances due to complex 3D motion, thereby enhancing fish tracking accuracy for
ecological studies. To fully answer this question we first take a look at the sub-questions. 
\par Single-View multi-objects trackers such as YOLOv8 with the ByteTrack adaption can to a certain extent concurrently track multiple fish entities on their own when trained on datasets containing fish entities. They however are not able to fully track all fish entities correctly for the entire duration of the videos. To enhance their performance post-re-identification processes are necessary, even after this tho they still cannot manage to fully track all fish entities. They can however track some fish entities for the entire duration of the videos and almost half for a limited duration of the videos. 
\par YOLOv8 as a single-view multi-object tracker can output data that can facilitate pattern recognition in fish locomotion. Our YOLOv8 model is able to output x and y coordinates for fish entities per frame allowing patterns to be found in the 2D movements of the fish entities.
\par Using the proposed framework it is possible to leverage dual video inputs to generate three-dimensional representations of fish entities. For certain videos, the framework is able to generate these 3D representations of fish entities for the majority of the fish. This is not possible in all videos tho due to our model not being able to detect and track all fish entities in all videos and our stereo-matching process not being able to properly match the fish entities either. However, when using manual matches we are able to create 3D representations of a limited set of detected fish entities. 
\par The incorporation of a 3D perspective creates new and more accurate tracking data containing 3D coordinates when compared to the 2D coordinate output of the non-leveraged models. The addition of a depth coordinate adds more data and makes pattern recognition more accurate as the speed of fish movements becomes more realistic, their depth changes are added and their grouping becomes more realistic. 
\par This all leads to our final conclusion that for a select set of videos our proposed multi-view MOT framework using single-view multi-object trackers can be adapted to overcome the challenges of fish tracking in datasets characterized by small and visually similar fish with rapidly changing appearances due to complex 3D motion and thereby enhancing fish tracking accuracy for ecological studies. When looking at all videos it is still true to a certain extent as the output does facilitate enhanced tracking accuracy when compared to the single-view models output and provides more data to be used for ecological studies. This research has thereby proven that these models can be used on underwater videos and can be dual leveraged which was before this research unknown. It adds to the vast research done on object detection and tracking by adding to the missing gap on underwater performance and animal, more specifically fish detection and tracking. 
\par For future work more research should be done into what other single-view models can be used for similar applications and how underwater footage can be made easier to detect and track objects in. The impact of applications such as sea-Thru \cite{akkaynak2019sea} that aim to enhance color contrast and "remove" the water from underwater footage could be studied to see if and how it improves the results of our framework. 


\newpage
\onecolumn
\onecolumn
\section{Appendix A: Most Common Matches DataFrame Example}
\begin{figure}[H]
    \centering
    \includegraphics[width=2cm,height=3cm]{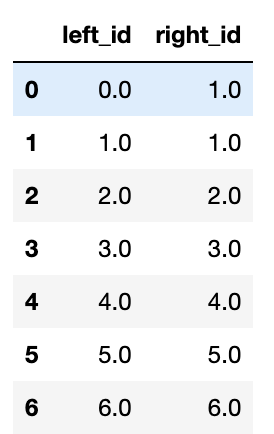}
    \caption{Stereo Matching DataFrame Video 33}
    \label{fig:ReID}
\end{figure}
\section{Appendix B: Mot Data Cleaning Example}
\begin{figure}[H]
    \centering
    \includegraphics[width=4cm]{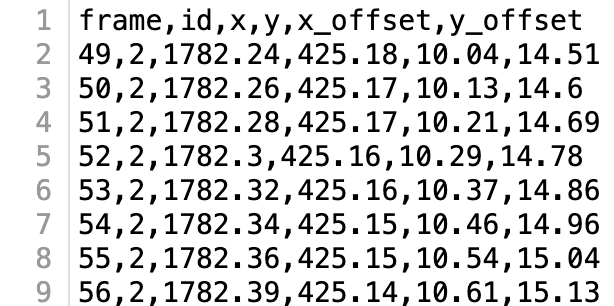}
    \includegraphics[width=4cm]{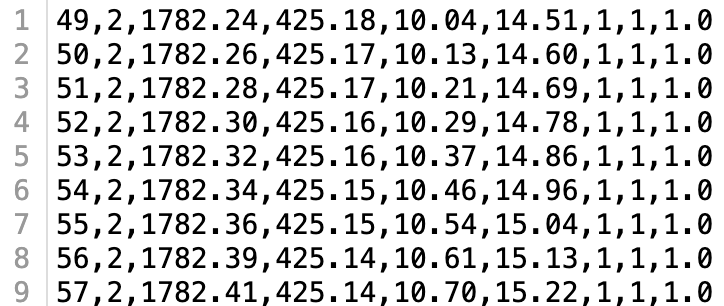}
    \caption{Mot Data Cleaning Before and After}
\end{figure}
\section{Appendix C: Visualization Output}
\subsection{Fish 0 Trajectory}
In Figure 12 we show the 3D trajectory of A fish entity with ID 0 from video 129. 
\begin{figure}[H]
    \centering
    \begin{tabular}{cc}
        \includegraphics[width=8cm]{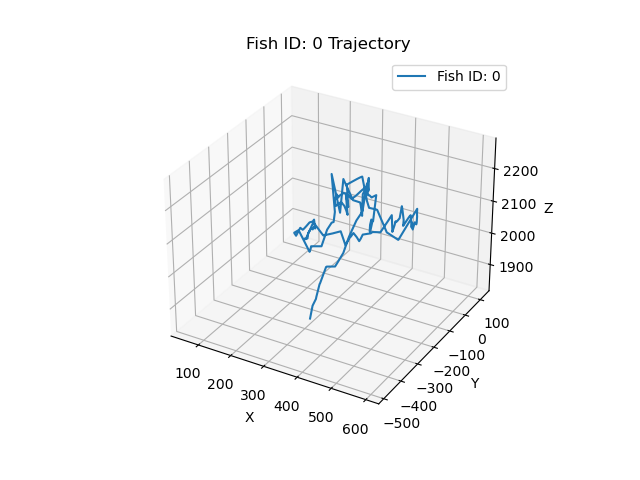}
    \end{tabular}
    \caption{Fish 0 Trajectory Video 129}
\end{figure}
\subsection{Fish Speed}
In Figure 13 we show the speed over time of A fish entity with ID 0 from video 129. 
\begin{figure}[H]
    \centering
    \begin{tabular}{cc}
        \includegraphics[width=6cm]{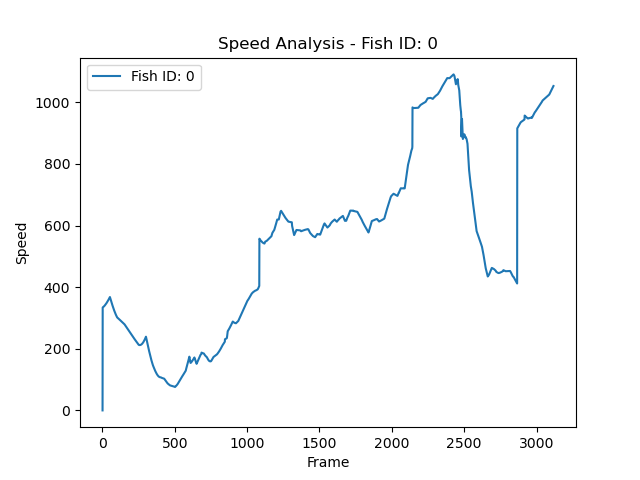} 
    \end{tabular}
    \caption{Fish 0 Speed Video 129}
\end{figure}
\subsection{Fish Acceleration}
In Figure 14 we show the acceleration over time of A fish entity with ID 0 from video 129.
\begin{figure}[H]
    \centering
    \begin{tabular}{cc}
        \includegraphics[width=6cm]{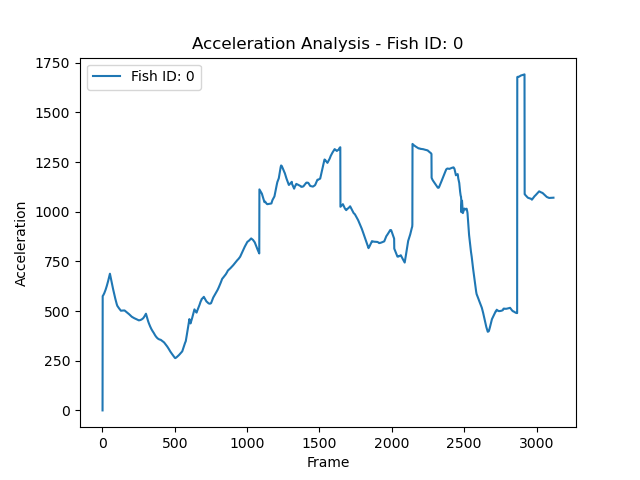} 
    \end{tabular}
    \caption{Fish 0 Acceleration Video 129}
\end{figure}
\subsection{Fish Path Lengths}
In Figure 15 we show the path lengths of fish entities with their corresponding ID from video 129. 
\begin{figure}[H]
    \centering
    \begin{tabular}{cc}
        \includegraphics[width=6cm]{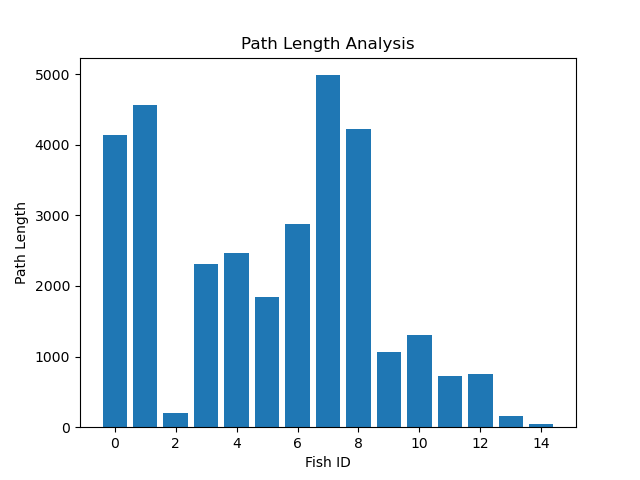} 
    \end{tabular}
    \caption{Fish Path lengths Video 129}
\end{figure}
\subsection{Fish Spatial Distribution}
In Figure 16 we show the spatial distributions of fish entities with their corresponding ID from video 129. 
\begin{figure}[H]
    \centering
    \begin{tabular}{cc}
        \includegraphics[width=6cm]{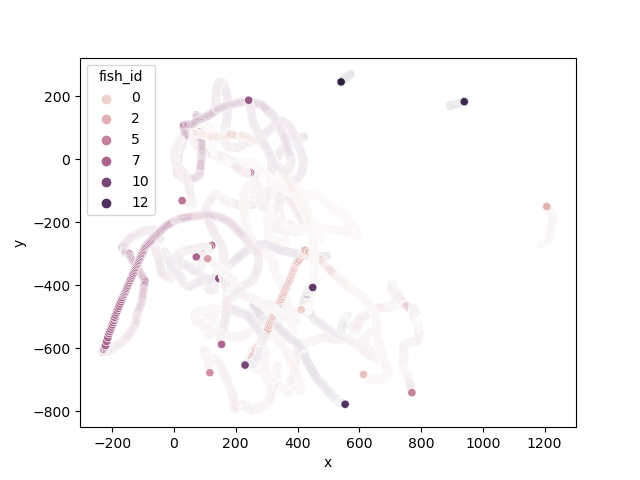} 
    \end{tabular}
    \caption{Fish Spatial Distributions Video 129}
\end{figure}
\subsection{Fish Density Map}
In Figure 17 we show the Density Map of fish entities with their corresponding ID from video 129. 
\begin{figure}[H]
    \centering
    \begin{tabular}{cc}
        \includegraphics[width=6cm]{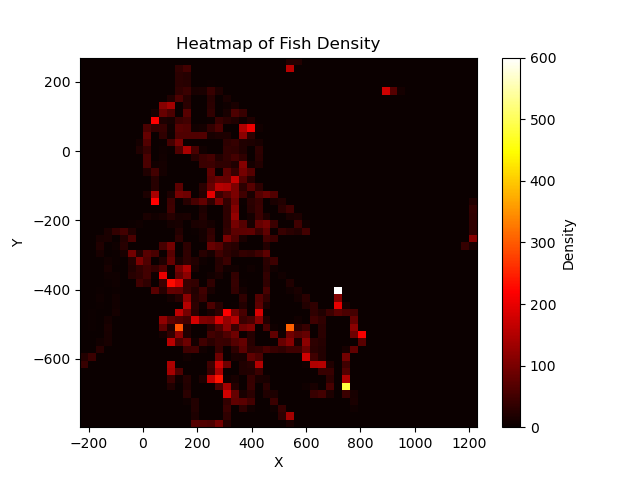} 
    \end{tabular}
    \caption{Fish Density Map Video 129}
\end{figure}
\subsection{Fish Temporal Patterns}
In Figure 18 we show the average temporal patterns of fish entities from video 129. 
\begin{figure}[H]
    \centering
    \begin{tabular}{cc}
        \includegraphics[width=6cm]{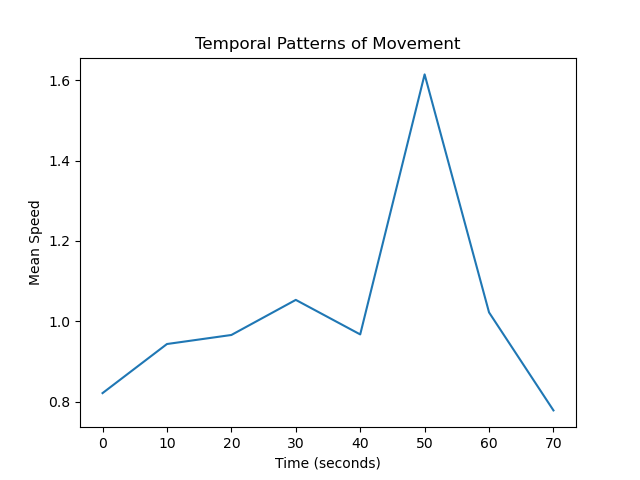} 
    \end{tabular}
    \caption{Average Fish Temporal Patterns Video 129}
\end{figure}
\subsection{Fish Depth}
In Figure 19 we show the depth over time of all fish entities with their corresponding ID from video 129.  
\begin{figure}[H]
    \centering
    \begin{tabular}{cc}
        \includegraphics[width=6cm]{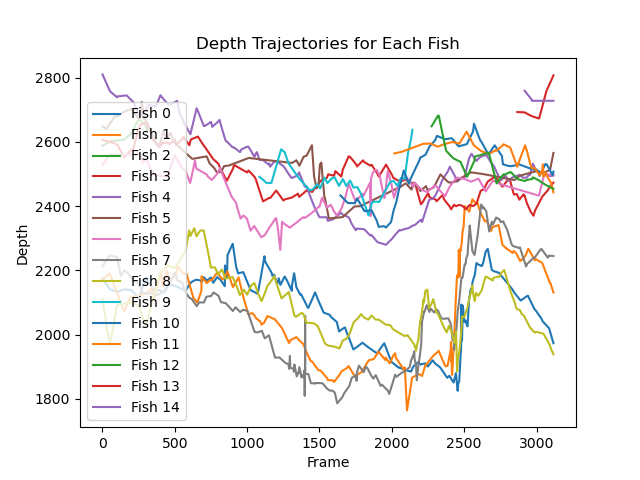} 
    \end{tabular}
    \caption{Fish Depths Video 129}
\end{figure}
\section{Appendix D:}
\begin{figure}[H]
    \centering
    \includegraphics[width=3cm, height=2.5cm]{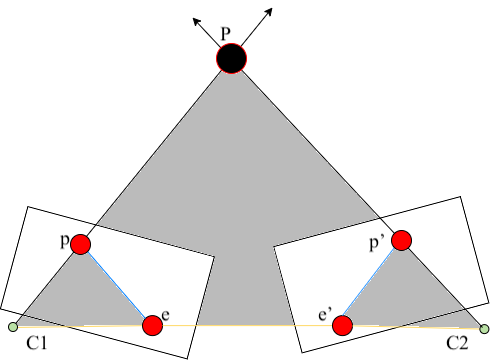}
    \small
    \caption{3D Coordinate Construction Using Epipolar Geometry with the gray region being the epipolar plane, the orange line the baseline, the blue lines the epipolar lines, 3D point P whose projection in each image is located at p and p', the camera centers (C1 \& C2) and the location of where the baseline intersects the two images planes e and e'. }
    \label{fig:ReID}
\end{figure}
\newpage
\section{Appendix E:}
\subsection{Fish Training Pipeline}
\begin{figure}[H]
    \centering
    \includegraphics[width=10cm, height=2.5cm]{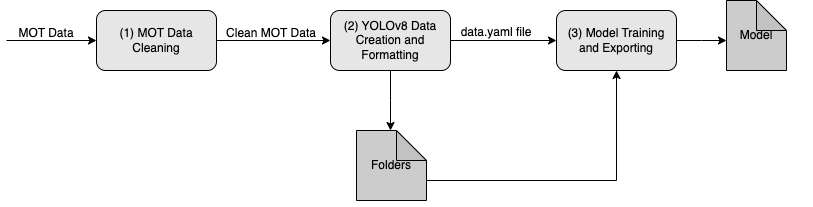}
    \small
    \caption{Training Pipeline Visualization}
    \label{fig:ReID}
\end{figure}
\subsection{Fish Tracking Pipeline}
\begin{figure}[H]
    \centering
    \includegraphics[width=19cm, height=5.5cm]{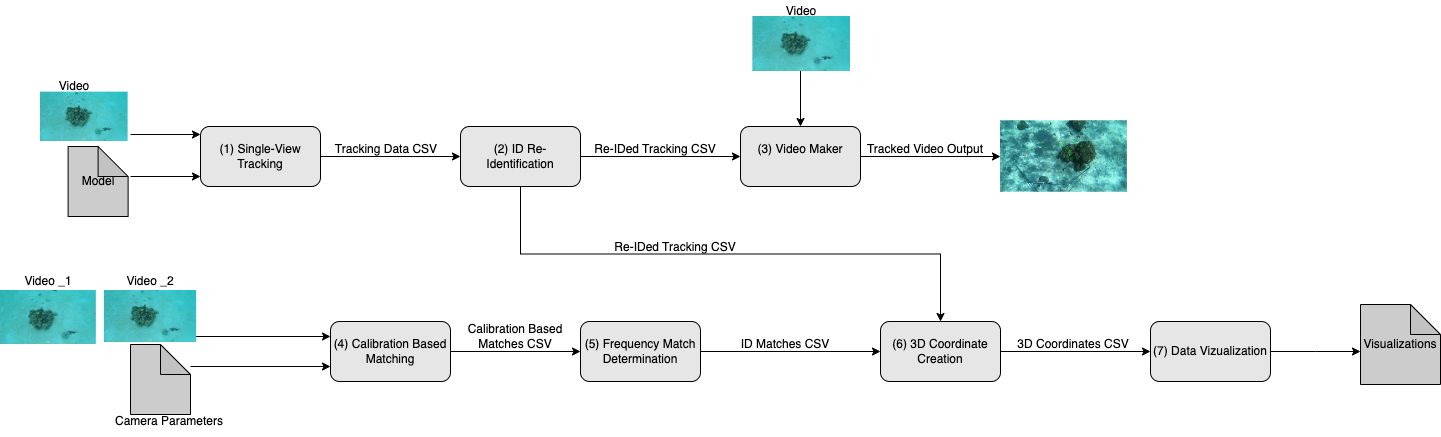}
    \small
    \caption{Training Pipeline Visualization}
    \label{fig:ReID}
\end{figure}
\section{Appendix F: Full Evaluation Metrics Table}

\end{document}